# An Evaluation of the Pedagogical Soundness and Usability of AI-Generated Lesson Plans Across Different Models and Prompt Frameworks in High-School Physics

Xincheng Liu[1] [a]

[a] The ISF Academy, 1 Kong Sin Wan Road, Pokfulam, Hong Kong

[a] charlesxliu25@gmail.com

**Abstract**

This study evaluates the pedagogical soundness and usability of AI-generated lesson plans across five leading large language models—ChatGPT (GPT-5), Claude Sonnet 4.5, Gemini 2.5 Flash, DeepSeek V3.2, and Grok 4—using three structured prompt frameworks: TAG (Task, Audience, Goal), RACE (Role, Audience, Context, Execution), and COSTAR (Context, Objective, Style, Tone, Audience, Response Format).

Fifteen lesson plans were generated for a single high-school physics topic, The Electromagnetic Spectrum. The lesson plans were analyzed through four automated computational metrics: (1) readability and linguistic complexity, (2) factual accuracy and hallucination detection, (3) standards and curriculum alignment, and (4) cognitive demand of learning objectives.

Results indicate that model selection exerted the strongest influence on linguistic accessibility, with DeepSeek producing the most readable teaching plan (FKGL = 8.64) and Claude generating the densest language (FKGL = 19.89).

The prompt framework structure most strongly affected the factual accuracy and pedagogical completeness, with the RACE framework yielding the lowest hallucination index and the highest incidental alignment with NGSS curriculum standards. Across all models, the learning objectives in the fifteen lesson plans clustered at the Remember and Understand tiers of Bloom's taxonomy. There were limited higher-order verbs in the learning objectives extracted.

Overall, the findings suggest that readability is significantly governed by model design, while instructional reliability and curricular alignment depend more on the prompt framework. The most effective configuration for lesson plans identified in the results was to combine a readability-optimized model with the RACE framework and an explicit checklist of physics concepts, curriculum standards, and higher-order objectives.

---

[1] Permanent Address: Tower 1 6A, 42 Kotewall Road, Mid-Levels, Hong Kong

## 1. Introduction

Since the release of ChatGPT and other generative AI tools, educators have increasingly explored their potential applications in the education field. Artificial Intelligence has opened up new possibilities for teaching and learning, both in K-12 and higher education (Dornburg & Davin, 2025; Fan et al., 2024).

Among the many uses for AI in education, one possibility includes using AI to streamline lesson planning. This is particularly attractive as it allows educators to save time and generate new ideas for their lessons. By inputting a simple prompt, AI tools can generate fully formatted lesson plans within seconds.

However, while AI-generated lesson plans are extremely convenient, concerns have arisen about their pedagogical quality (Karaman & Göksu, 2024; Powell & Courchesne, 2024). These concerns primarily pertain to their accuracy, their adaptability to student needs, and overall classroom feasibility. While previous studies have praised AI's ability to produce logically organized and human-like lesson plans efficiently, many have emphasized that human guidance is essential to ensure effectiveness. Moreover, most prior evaluations and studies of AI lesson planning have focused on single AI models (Baytak, 2024), leaving a gap in cross-model evaluations of AI-generated lesson plans.

Additionally, a small but growing amount of research has explored how the structure and detail of prompts can influence the quality of AI outputs. Prompt engineering frameworks such as TAG (Task, Audience, Goal), RACE (Role, Audience, Context, Execution), and COSTAR (Context, Objective, Style, Tone, Audience, Response Format) have been developed and tested to optimize the quality of AI outputs. These frameworks have been applied and investigated in various domains; however, little to no research has examined how these prompt frameworks specifically influence the quality of AI-generated lesson plans. Understanding this relationship is essential for educators to effectively harness AI tools to generate lesson plans.

To address these research gaps, this study presents an evaluation of lesson plans generated by five leading AI chatbot models—ChatGPT (GPT-5), Claude Sonnet 4.5, Gemini 2.5 Flash, DeepSeek V4.2, and Grok 4—for a single high-school physics topic. Each chatbot was asked to generate lesson plans for the topic of The Electromagnetic Spectrum using three different prompt frameworks: TAG, RACE, and COSTAR. This research design allows an assessment not only of how different AI models differ in pedagogical performance but also of how different prompt frameworks influence the output quality of AI-generated teaching plans.

## 2. Objective

The main objective of this study is to evaluate the accuracy, quality, and feasibility of lesson plans generated by the five leading free-access AI chatbots. This investigates how suitable it is to utilize Artificial Intelligence to create lesson plans for the classroom. The objectives of the study can be categorized into 2 specific objectives:

SO1. Evaluate the pedagogical quality of AI-generated lesson plans produced by five leading generative AI models (ChatGPT, Claude, Gemini, Grok, and DeepSeek).

SO2. Investigate the impact of prompt framework design by analyzing how TAG, RACE, and COSTAR prompt frameworks influence the instructional quality, structure, and creativity of AI-generated lesson plans.

By pursuing these objectives, this study aims to provide evidence-based insights into how both AI chatbot choice and prompt framework influence the quality of AI-generated lesson plans, offering guidance for educators.

## 3. Literature Review

The use of Large Language Models (LLMs) for lesson planning has gained some significant attention and traction among educators. Following this trend, education researchers have begun to systematically evaluate the quality, usability, and effectiveness of AI-generated lesson plans. This meta-analysis will synthesize current research that directly assesses AI-generated lesson plans across multiple dimensions.

### 3.1 Pedagogical Quality and Student Outcomes

Some controlled studies have found that AI-generated lesson plans can match or even exceed the effectiveness of traditionally developed plans when measured by student achievement through test scores.

Karaman and Göksu (2024) conducted a pretest–posttest quasi-experimental study involving 39 third-grade students, with 24 students in an experimental group and 15 students in a control group (Karaman & Göksu, 2024). The experimental group was taught using lesson plans generated entirely by ChatGPT, while the control group received lessons based on teacher-developed plans. The students took a multiple-choice test of 25 questions at the start and end of the experiment. Over the course of a five-week mathematics unit, both groups demonstrated significant gains, with very large effect sizes (Cohen's d ≈ 1.25–1.27). The ChatGPT group improved from a mean pre-test score of 38.4 to a post-test score of 61.2 ($t(23) = 6.47$, $p < .001$, $d = 1.268$), while the teacher-plan group improved from 34.0 to 55.3 ($t(14) = 5.54$, $p < .001$, $d = 1.250$). Importantly, there was no statistically significant difference between the groups' final scores ($t = 0.851$, $p = 0.400$), indicating that AI-generated plans were at least as effective as teacher-created plans in promoting learning outcomes.

In a similar study, Fan et al. (2024) developed a tool called *LessonPlanner,* which embeds Gagné's instructional design principles into an AI-driven lesson planning interface (Fan et al., 2024). By including 12 teachers as evaluators, these expert evaluators scored LessonPlanner plans higher in areas such as instructional clarity, alignment with learning objectives, integration of theory and practice, and feasibility. In addition to improved quality, teachers reported reduced planning time and cognitive load.

Together, these findings suggest that AI-generated lesson plans, especially when supported by structured frameworks or design scaffolds, can be pedagogically robust and instructionally effective. However, these benefits depend heavily on user input, prompt structure, and post-generation editing.

## 3.2 Structural Consistency and Prompt Variability

Research on Artificial Intelligence over the past two years has also examined the field of prompt engineering.

As for prompt variability in AI-generated lesson plans, Dornburg and Davin from Cambridge University conducted a detailed evaluation of ChatGPT-generated foreign-language lesson plans under increasingly specific prompts(Dornburg & Davin, 2025). Five prompts were designed to include increasing amounts of context and structure. The first few prompts sometimes omitted essential components like formative assessments, context, or learning objective alignment. However, in the latter prompts, when rubrics or stepwise instructions were embedded in prompts, outputs were significantly more complete and coherent. This shows that AI models do not reliably infer pedagogical priorities unless explicitly guided, making the human role in creating and framing prompts crucial towards generating an effective teaching plan.

In the study, Dornburg and Davin also observed that AI-generated plans often reflected superficial coherence but lacked deeper instructional logic. That is, sequencing activities didn't gradually increase in cognitive complexity. This supports the notion that AI lesson planning largely requires human guidance and scaffolding to achieve pedagogically meaningful and effective teaching plans (Dornburg & Davin, 2024, 2025). While LLMs do have the ability to generate grammatically coherent and basic plans, they fail to implement more deeply productive scaffolding. This shows that AI models are not able to infer pedagogical priorities unless being guided explicitly, hence making the design of prompts crucial.

Taken together, Dornburg and Davin's research consolidates the notion that prompt specificity, clarity, and embedded evaluation criteria directly affect the quality, consistency, and pedagogical fidelity of AI-generated lesson plans.

## 3.3 Instructional Characteristics and Alignment with Standards

While AI-generated lesson plans are shown to be able to achieve a basic level of structural coherence, studies have noted the fact that these lesson plans contain limitations in curricular alignment, differentiation, and connections to real-life implications and adaptation to local contexts.

Baytak (2024) analyzed a content analysis of 18 lesson plans generated by ChatGPT and Google Gemini across the 7th-grade mathematics, science, literature, and social studies classes(Baytak, 2024). The AI outputs were observed to generally produce clear objectives, logically sequenced activities, and basic assessment tasks, thus allowing them to be sufficient to be used in lessons. However, it was noted that the learning objectives from the generated lesson plans were narrower than the learning objectives of the New Jersey Student Learning Standards, hence showing that they rarely included links to national, state, or international curricular standards. Further, it was pointed out that the generated lesson plans by ChatGPT and Gemini did not meaningfully integrate technology tools or strategies, but rather included simple materials such as "whiteboard or projector", "markers or pens", and "handouts or articles".

Kahraman and Kıyıcı (2025) evaluated the effectiveness of artificial intelligence in generating middle school science lesson plans against Turkey's national science curriculum (Kahraman and Kıyıcı, 2025). While the objectives of the class and real-world context were addressed, the generated plans lacked formative assessment loops, reflection, and differentiation strategies for students with varying needs and abilities. Similarly, Lee and Zhai (2024) found that pre-service teachers' AI-assisted lesson plans were partially aligned with curricular standards and required adaptation to fit the local classroom needs (Lee & Zhai, 2024).

Powell and Courchesne (2024) investigated the opportunities and risks involved in utilizing ChatGPT to generate first-grade science lesson plans. They found an additional risk that the lesson plans included factual errors and fabricated resources (Powell & Courchesne, 2024).

Overall, studies indicate that AI-generated lesson plans excel at coherence but consistently fall short in deeper pedagogical aspects (Baytak, 2024; Kahraman & Kıyıcı, 2025; Lee & Zhai, 2024; Powell & Courchesne, 2024). These include alignment with curriculum, differentiation, formative assessment, and context-sensitive adaptations. These findings suggest that AI tools are most effective when used in partnership with experienced educators who can validate and refine outputs.

**3.4 Summary of Findings**
Across studies, several common patterns emerge:
1. Comparable effectiveness: AI-generated lesson plans are capable of producing learning outcomes comparable to human-designed plans when combined with structured prompts and scaffolds
2. Importance of prompting: Embedding instructional frameworks, rubrics, or checklists consistently enhances output quality and reduces variability
3. Curricula alignment gaps: AI plans often lack explicit references to standards, differentiation strategies, technology integration, and feedback
4. Human intervention: Across all studies, teacher intervention has been stated to be necessary to ensure content accuracy, pedagogical soundness, and suitability for diverse learners

In conclusion, research shows that AI lesson planning shows substantial promise. However, its effectiveness is contingent upon structured prompt design, human guidance, and meticulous curricular adaptation.

The literature analysis identifies a clear gap in the research of AI lesson planning: cross-model comparative analyses examining how AI system choice and prompt framework influence the quality of lesson plans. The present study addresses this need by evaluating five leading LLMs (ChatGPT, Claude, Gemini, DeepSeek, Grok) using three prompt frameworks (TAG, RACE, and COSTAR). Hence, this provides the first systematic multi-dimensional assessment of AI-generated lesson plans.

## 4. Methodology

The methodology implements a single, fully automated evaluation to analyze the pedagogical quality of AI-generated lesson plans produced by different AI models under different prompt frameworks. The automated evaluation applies four predefined metrics to each lesson plan: readability and linguistic complexity, factual accuracy and hallucination detection, standards alignment, and the cognitive level of learning objectives. The aim is to determine how effectively each AI model can generate structured, pedagogically sound lesson plans that align with established teaching and learning principles.

Five of the most popular AI models were selected: ChatGPT (GPT-5), Claude Sonnet 4.5, Gemini 2.5 Flash, DeepSeek V3.2, and Grok 4. Each model was prompted to generate lesson plans for the subject physics using three distinct prompt frameworks: TAG (Task–Audience–Goal), RACE (Role–Action–Context–Execute), and COSTAR (Context–Objective–Style–Tone–Audience–Response Format). This factorial design (5 models × 3 prompts) results in a total of fifteen AI-generated lesson plans. For comparability, all lesson plans were generated under standardized conditions in June 2025.

### 4.1 Lesson Topic

The physics topic selected for the prompt was The Electromagnetic Spectrum. This topic was chosen as it represents a core high-school physics topic that integrates conceptual understanding, mathematics, and real-world application. The prompts instructed each AI model to produce a complete 60-minute lesson plan for a mixed-ability classroom of Grade 9 to 12 students. The generated lesson plans were expected to include clear learning objectives, differentiation strategies, assessment methods, and a closing reflection activity.

### 4.2 Prompt Frameworks and Input Design

Three widely used prompt frameworks were applied to examine how structured inputs influence the quality of AI-generated lesson plans. To ensure impartiality, the amount of detail, contextual guidance, and instructional requirements was kept identical across all prompts. Each version of the prompt contained the same information about lesson duration (60 minutes), class type (mixed-ability students in Grades 9-12), required lesson components, and topic. The only variable manipulated was the organizational structure of the framework itself. This design isolates the effect of framework structure on the output quality of AI-generated lesson plans rather than differences in prompt specificity.

The prompts can be seen below:

TAG (Task–Audience–Goal):
- Task: Generate a complete 60-minute lesson plan on The Electromagnetic Spectrum.
- Audience: Mixed-ability high-school students (Grades 9–12).
- Goal: Ensure learning objectives are clear, activities are engaging and age-appropriate, and the plan includes differentiation strategies, hands-on demonstrations, formative assessment, and a closing reflection.

RACE (Role–Action–Context–Execute):

- Role: You are a high-school physics teacher preparing a 60-minute lesson.
- Action: Design a detailed lesson plan on The Electromagnetic Spectrum.
- Context: The class consists of mixed-ability students in Grades 9–12. Lessons should include clear learning objectives, engaging and age-appropriate activities, differentiation strategies, hands-on demonstrations, formative assessment, and a closing reflection.
- Execute: Present the plan in a structured format suitable for classroom use, including objectives, materials, procedures, and assessment methods.

COSTAR (Context–Objective–Style–Tone–Audience–Response Format):
- Context: You are a high-school physics teacher teaching a mixed-ability high-school physics class (Grades 9–12) learning about The Electromagnetic Spectrum.
- Objective: Create a 60-minute lesson plan where learning objectives are clear, activities are engaging and age-appropriate, and the lesson includes differentiation strategies, hands-on demonstrations, formative assessment, and a closing reflection
- Style: Structured, practical, and easy for a teacher to implement.
- Tone: Professional, supportive, and student-centred.
- Audience: Grades 9–12 students with varying levels of prior science knowledge.
- Response Format: Provide a complete lesson plan including objectives, materials, step-by-step activities, assessment methods, and a short teacher reflection section.

By maintaining uniform content density and expectations across all the prompts, this setup ensures that any observed differences in generated teaching plans stem solely from differences in prompt framework structure.

### 4.3 Data Collection
Each AI model generated one lesson plan per prompt, resulting in a total of fifteen lesson plans. All outputs of lesson plans were anonymized and stored in text format (.txt) for computational analysis.

After export, all files were anonymized and normalized to support automated processing. No content edits were made.

### 4.4 Automated Evaluation Pipeline
A fully automated evaluation pipeline was implemented to quantify the pedagogical and linguistic characteristics of each lesson plan. A total of four computational metrics were applied:
1. Readability and Linguistic Complexity
2. Factual Accuracy and Hallucination Detection
3. Standards and Curriculum Alignment
4. Cognitive Level of Learning Objectives

All computational analyses were executed in Python 3.10. Established natural-language and data-processing libraries, including pandas, nltk, scikit-learn, and sentence-transformers, were used to compute quantitative data across all model-framework combinations.

### 4.4.1 Readability and Linguistic Complexity

The first metric is the readability and linguistic complexity metric. This metric evaluates the accessibility and linguistic appropriateness of each lesson plan. Readability is a critical determinant of whether AI-generated lesson plans can be effectively understood and implemented by teachers and students. In education materials, text that is too complex or overly simplified can hamper comprehension or reduce conceptual depth. To quantify these linguistic characteristics, several readability and linguistic complexity measures were applied.

The Flesch-Kincaid Grade Level (FKGL) index estimates the U.S. school grade level required to comprehend a given text (Flesch, 1948). It is calculated by:

$$FKGL = 0.39\left(\frac{total\ words}{total\ sentences}\right) + 11.8\left(\frac{total\ syllables}{total\ words}\right) - 15.59$$

Higher FKGL values indicate a more linguistically demanding text with more polysyllabic words and longer sentences. The FKGL value for students in grades 9-12 ranges from 9-12. The Flesch Reading Ease (FRE) score, in contrast, inversely measures readability difficulty. It is given by the formula:

$$FRE = 206.835 - 1.015\left(\frac{total\ words}{total\ sentences}\right) - 84.6\left(\frac{total\ syllables}{total\ words}\right)$$

Scores between 60 and 70 generally correspond to plain English that is comprehensible to high-school students, while scores below 30 indicate highly academic texts (Flesch, 1948).

The Gunning Fog Index (FOG) provides an additional measure of readability. It incorporates the percentage of complex words (words with three or more syllables) into analyses. The calculation is expressed as:

$$FOG = 0.4\left[\left(\frac{total\ words}{total\ sentences}\right) + 100\left(\frac{complex\ words}{total\ words}\right)\right]$$

Gunning Fog Index values near 10 to 12 correspond to text suitable for grade 9-12 audiences, while values above 16 indicate writing similar to an academic paper.

Furthermore, to capture the lexical diversity of the lesson plans, the Type–Token Ratio (TTR) is computed. The TTR is defined as the number of unique words divided by the total number of words in a text. Higher TTR values indicate a more complex and richer vocabulary use.

Finally, basic measures of syntactic cohesion were calculated through values such as average sentence length. The average sentence length was calculated by obtaining the mean number of words per sentence. Longer average sentence lengths tend to indicate greater syntactic complexity and more difficult readability.

Each readability and linguistic index was calculated following all lesson plans being converted to text files (.txt). These linguistic metrics were then aggregated across all teaching plans for lessons and frameworks to create a composite linguistic profile for each model and framework. This serves as an indicator of how effectively AI-generated lesson plans balance clarity and sophistication.

Within educational contexts, readability directly impacts pedagogical usability. Hence, the linguistic profile will also provide insight into the feasibility of such lesson plans for the target audience.

### 4.4.2 Factual Accuracy and Hallucination Detection

The second metric is the factual accuracy and hallucination detection metric. It examines the factual reliability of the AI-generated lesson plans. As physics instruction and lessons depend on accurate communication of mathematical and conceptual relationships, any minor error can distort scientific understanding. A rule-based verification pipeline is implemented to identify and quantify factual inconsistencies or hallucinated content across all teaching plans.

Omissions in the lesson plans of key information and conceptual relationships were treated as errors. A plan scored an omission error whenever a targeted relationship or constant was not present.

In the first stage of the verification process, the pipeline extracted all numerical constants, equations, and domain-specific entities from the text. Regular expressions were used to detect canonical expressions about The Electromagnetic Spectrum, such as $c = \lambda f$, $E = hf$, and $E = \frac{hc}{\lambda}$. Constants, including the speed of light, Planck's constant, and frequency or wavelength units, were also extracted to validate.

In the second stage, these extracted entities were cross-validated against the values from verified databases. Deviations greater than ten percent from the literature values were categorized as minor errors, while categorical inversions or conceptual errors were classified as major errors. Omissions of any target item were counted as minor errors.

Additionally, the pipeline identified fabricated information from the generated lesson plans, including made-up experiments or nonexistent concepts. Each plan's cumulative factual reliability was quantified using a developed Hallucination Index (HI). The HI is defined as:

$$HI = (2 \times major\ errors) + (1 \times minor\ errors)$$

Lower HI values indicate higher factual accuracy. This numerical index provides a consistent measure of factual soundness. Within the educational domain, this precision is especially crucial, as factual errors lead to misconceptions that hinder conceptual learning.

### 4.4.3 Standards and Curriculum Alignment

The third metric is the standards and curriculum alignment metric. This metric assessed the degree of alignment of each AI-generated lesson plan with established educational standards. In particular, the Next Generation Science Standards (NGSS) for high-school physics was chosen on the topic of The Electromagnetic Spectrum, which was the HS-PS4 cluster.

As none of the prompts explicitly instructed the AI models to follow a specific curriculum, this analysis captures how spontaneously the models generated learning objectives that resemble those of recognized physics standards.

To measure alignment, each lesson plan was first scanned to extract sentences that expressed learning objectives. These objectives were then converted into semantic embeddings using the Sentence-BERT language model with the sentence-transformers library. The same process was applied to the five official standards for the NGSS HS-PS4 cluster.

Next, the study calculated the cosine similarity between both texts. Cosine similarity measures how similar two texts are in terms of meaning. The formula is given by:

$$S = \frac{A \cdot B}{||A|| \, ||B||}$$

A and B in this case are vector representations of the two learning objective texts. The cosine similarity calculation gives a score on a scale from 0 to 1. A score of 0 indicates no alignment, while a score of 1 means that both texts are identical in meaning.

Scores below 0.20 generally indicate weak alignment, while scores between 0.20 and 0.35 suggest moderate alignment. Further, scores above 0.35 reflect strong conceptual alignment. This method provides a clear way to determine how closely AI-generated lesson plans reflect the curriculum of formal physics education.

**4.4.4 Cognitive Level of Learning Objectives**
The fourth metric is the cognitive level of the learning objectives metric. This metric assessed the cognitive rigour of the generated lesson plans by classifying the learning objectives of each lesson plan according to Bloom's Revised Taxonomy (Anderson & Krathwohl, 2001).

Bloom's Taxonomy is a widely used framework that classifies learning objectives by cognitive complexity. The six levels of cognitive complexity range from recall to creative problem solving. The taxonomy is widely used in instructional design to evaluate the intellectual complexity and demand of educational materials.

Each learning objective from the lesson plans was first parsed to extract the principal verb using dependency parsing. The extracted verb was then matched against a curated vocabulary list of over 250 pedagogical verbs, each classified according to Bloom's six levels of thinking: Remember (1), Understand (2), Apply (3), Analyze (4), Evaluate (5), and Create (6).

Each verb extracted from the teaching plan was then assigned a numerical score corresponding to its taxonomy level. The mean score was calculated across all objectives in the lesson plan to compute a Cognitive Demand Index (CDI). This index provided a continuous measure of cognitive rigor, ranging from 1 to 6.

A lower CDI (1-2) represents a focus merely on factual knowledge and comprehension tasks. A moderate CDI (3-4) suggests that the plan moves beyond factual knowledge and encourages analytical and procedural thinking. Higher CDI values (5-6) signify engagement with evaluative judgement and inquiry-based learning, representations of advanced scientific reasoning. This classification of a Cognitive Demand Index allowed the study to identify whether AI lesson plans primarily produced surface-level learning outcomes or successfully incorporated higher-order thinking that aligned with more contemporary pedagogical goals in STEM education.

### 4.4.5 Summary
Together, these four metrics form an analytical framework for evaluating the pedagogical soundness of AI-generated lesson plans. This methodological approach enables objective, quantitative, and replicable assessment that allows for direct comparison across models and prompt frameworks.

### 4.5 Statistical Aggregation
All automated metrics were first computed for each plan and then aggregated to the model and frameworks. For the model-level aggregation, each model's three outputs generated under TAG, RACE, and COSTAR were averaged to obtain a single composite score per metric. For the framework-level aggregation, scores were averaged across all five models. All reported values, therefore, represent mean performance metrics. This provides a stable basis for cross-model and cross-framework comparison in Section 5.

## 5. Results
All fifteen lesson plans were generated across the five AI models and underwent computational analysis (See Appendix A).

This section presents quantitative findings from the automated evaluation across readability and linguistic complexity, factual accuracy and hallucination detection, standards and curriculum alignment, and the cognitive demand of learning objectives.

### 5.1 Automated Evaluation Results
Below are all the evaluated results for all metrics across frameworks and models.

| **Framework** | TTR | FRE | FKGL | FOG | HI |
|---|---|---|---|---|---|
| **COSTAR** | 0.43 | 38.06 | 12.43 | 15.53 | 2.90 |
| **RACE** | 0.48 | 34.08 | 12.76 | 15.89 | 2.20 |
| **TAG** | 0.45 | 35.37 | 13.05 | 16.06 | 2.30 |

Table 1. Readability and Hallucination Metrics Averaged by Framework

| Framework | HS-PS4-1 | HS-PS4-3 | HS-PS4-4 | HS-PS4-5 | NGSS Mean |
|---|---|---|---|---|---|
| **COSTAR** | 0.148 | 0.045 | 0.052 | 0.035 | 0.070 |
| **RACE** | 0.139 | 0.074 | 0.074 | 0.042 | 0.082 |
| **TAG** | 0.168 | 0.057 | 0.051 | 0.046 | 0.081 |

Table 2. Curriculum Alignment Metrics Averaged by Framework

| Framework | Bloom - Remember | Bloom - Understand | Bloom - Apply | Bloom - Analyze | Bloom - Evaluate | Bloom - Create | CDI |
|---|---|---|---|---|---|---|---|
| **COSTAR** | 0.494 | 0.301 | 0.000 | 0.147 | 0.057 | 0.000 | 1.969 |
| **RACE** | 0.430 | 0.338 | 0.040 | 0.158 | 0.033 | 0.000 | 2.023 |
| **TAG** | 0.342 | 0.368 | 0.175 | 0.114 | 0.000 | 0.000 | 2.059 |

Table 3. Cognitive Metrics Averaged by Framework

| Model | TTR | FRE | FKGL | FOG | HI |
|---|---|---|---|---|---|
| **ChatGPT** | 0.46 | 41.73 | 10.86 | 14.04 | 2.50 |
| **Claude** | 0.47 | 9.89 | 19.89 | 23.29 | 2.67 |
| **DeepSeek** | 0.43 | 54.65 | 8.64 | 11.47 | 2.83 |
| **Gemini** | 0.45 | 43.44 | 11.03 | 14.20 | 2.33 |
| **Grok** | 0.47 | 29.47 | 13.32 | 16.14 | 2.00 |

Table 4. Readability and Hallucination Metrics Averaged by Model

| Model | HS-PS4-1 | HS-PS4-3 | HS-PS4-4 | HS-PS4-5 | NGSS Mean |
|---|---|---|---|---|---|
| **ChatGPT** | 0.130 | 0.042 | 0.043 | 0.038 | 0.063 |
| **Claude** | 0.190 | 0.084 | 0.107 | 0.058 | 0.110 |
| **DeepSeek** | 0.115 | 0.044 | 0.049 | 0.039 | 0.061 |
| **Gemini** | 0.142 | 0.033 | 0.043 | 0.027 | 0.061 |
| **Grok** | 0.183 | 0.091 | 0.052 | 0.043 | 0.092 |

Table 5. Curriculum Alignment Metrics Averaged by Model

| Model | Bloom - Remember | Bloom - Understand | Bloom - Apply | Bloom - Analyze | Bloom - Evaluate | Bloom - Create | CDI |
|---|---|---|---|---|---|---|---|
| **ChatGPT** | 0.340 | 0.387 | 0.067 | 0.056 | 0.151 | 0.000 | 2.294 |
| **Claude** | 0.312 | 0.401 | 0.000 | 0.287 | 0.000 | 0.000 | 2.262 |
| **DeepSeek** | 0.381 | 0.310 | 0.048 | 0.262 | 0.000 | 0.000 | 2.193 |
| **Gemini** | 0.600 | 0.222 | 0.178 | 0.000 | 0.000 | 0.000 | 1.578 |
| **Grok** | 0.479 | 0.360 | 0.067 | 0.095 | 0.000 | 0.000 | 1.780 |

Table 6. Cognitive Metrics Averaged by Model

### 5.2 Readability and Linguistic Complexity

According to the computational analysis presented in Section 5.1, the linguistic accessibility shows substantial variation by AI model and moderate variation by prompt framework.

Across all lesson plans, the Flesch-Kincaid Grade Level ranged from 7.63 to 21.01 (Claude, COSTAR), while Flesch Reading Ease (FRE) ranged from 6.91 to 58.44 across the same two lesson plans. These results suggest that, depending on model choice, AI systems can generate lesson plans ranging from high-school level readability to university-level academic readability and complexity.

When metrics are averaged by model, DeepSeek produced the most accessible lesson plan, with a mean FKGL of approximately 8.64 and a mean FRE of 54.65. Both of these values align well with the comprehension level of mixed-ability high-school students in grades 9-12. In contrast, Claude and Grok generated text that was considerably more complex, with mean FKGL values of 19.89 and 13.32. This is characterized by longer average sentence lengths and a higher proportion of polysyllabic words within the teaching plans. Their outputs often included subordinate clauses, increasing readability difficulty.

Gemini and ChatGPT occupied an intermediate position in terms of readability within the models, having mean FKGL values of 11.03 and 10.86, respectively. These values maintain a moderate readability suitable for students in grades 9-12.

Prompt framework effects on readability were present but insignificant compared to model effects. On average, COSTAR yielded the lowest mean FKGL (12.43). This is followed closely by RACE (12.76) and TAG (13.05).

These findings indicate that the AI model used is the primary determinant of readability, while framework structure plays a minor role in influencing readability. Models such as DeepSeek produce text that requires minimal to no editing for mixed-ability classrooms of students in grades 9-12, whereas Claude's dense academic register may be more suitable for teacher reference materials rather than direct teaching for student use.

Teachers who are aiming for more readable and balanced lesson plans should prioritize selecting a linguistically appropriate and accessible AI model.

### 5.3 Factual Accuracy and Hallucination Detection

Factual Accuracy and Hallucination Detection, defined by the Hallucination Index created varied to a similar degree between models and frameworks.

At the model level, Grok demonstrated the strongest factual reliability with a mean Hallucination Index of approximately 2.00. Its lesson plans consistently included the wave equation ($c = \lambda f$) and clear differentiation between the seven spectrum bands, but sometimes omitted the numerical constant for light speed and Planck's constant. ChatGPT and Gemini followed closely with Hallucination indices of 2.50 and 2.33. DeepSeek yielded the highest Hallucination Index value of 2.83, reflecting more frequent omissions and limited quantitative references. Overall, none of the lesson plans showcased any factual inaccuracies; however, several plans contained omissions of expected items, which is reflected in non-zero Hallucination Index scores.

Prompt framework effects on factual accuracy were more pronounced than on readability. The RACE prompt framework had the lowest mean Hallucination Index (2.0), followed by TAG (2.3) and COSTAR 2.9). This pattern suggests that either the procedural scaffolding of the RACE framework, particularly the "Role" and "Execute" components, encourages the inclusion of equations, constants, and ordered listings by framing the model's output as a professional teacher's deliverable, or the addition of "Style" and "Tone" of the COSTAR framework encouraged too much of a focus on overall response manner rather than specific content.

These results reveal that factual reliability is sensitive to both model choice and prompt framework. While Grok consistently produced the most accurate content, the RACE framework also offered a potential procedural advantage across models.

### 5.4 Standards Alignment

Although not explicitly instructed to do so, from the initial observation of the teaching plans, all three teaching plans generated by Claude Sonnet 4.5 automatically aligned lesson content with the Next Generation Science Standards (NGSS) of HS-PS4-1, HS-PS4-3, and HS-PS4-5 related to wave properties and electromagnetic radiation. This was explicitly stated in the teaching plans generated by Claude. Two teaching plans generated by Grok and one teaching plan generated by ChatGPT also followed the NGSS standards and explicitly indicated alignment.

At the model level, Claude displayed the highest average NGSS similarity with a similarity of 0.110, indicating that its outputs most closely approximated the structure and phrasing of curriculum-based objectives under the HS-PS4 cluster. Other models clustered near this value with small variances, with Grok also having a high similarity of 0.092 and ChatGPT having a similarity value of 0.063. This suggests overall moderate alignment, considering no specific instruction based on curriculum alignment was given.

Framework effects followed a similar hierarchy to that found in the factual analysis. RACE produced the strongest alignment with an NGSS mean of 0.082, followed by TAG with an NGSS mean of 0.081 and COSTAR with an NGSS mean of 0.070.

When looking at specific standards, alignment was strongest for HS-PS4-1—which addresses the relationships among frequency, wavelength, and speed—since nearly all plans included references to the mathematical relationship of $c = \lambda f$. Alignment with HS-PS4-3, HS-PS4-4, and HS-PS4-5, which separately focus on models of electromagnetic radiation, effects of different frequencies of electromagnetic radiation when absorbed by matter, and the application of waves in technological devices, was moderately low compared to alignment with HS-PS4-1. This could be explained as the latter three standards are not the main high-school physics topics within the topic of The Electromagnetic Spectrum.

The findings indicate that, in the absence of explicit standards in the prompt, models naturally approximate the foundational concepts of HS-PS4-1 but struggle with higher-level applications that involve cross-disciplinary reasoning. This can be observed most significantly in AI models Claude and Grok. Frameworks such as RACE improve incidental alignment by ensuring that objectives are expressed with procedural completeness. Ultimately, to achieve stronger curricular correspondence, teachers must incorporate explicit NGSS references and guidance so that the generated lesson plan follows the curriculum standards.

**5.5 Cognitive Demand of Objectives**
Bloom's taxonomy analysis revealed that most objectives in the AI-generated lesson plans were concentrated at the lower cognitive tiers.

Across all models, verbs associated with Remember (e.g., identify, list, describe) and Understand (e.g., explain, compare, discuss, interpret) accounted for the majority of occurrences. These two verbs represent the two lowest levels of cognitive demand in Bloom's taxonomy.

ChatGPT, Claude, and DeepSeek exhibited slightly higher proportions of Understand-level verbs (0.310–0.401), with overall higher values for the Cognitive Demand Index, at 2.294, 2.262, and 2.193. This suggests a stronger emphasis on conceptual comprehension. Gemini and Grok both leaned more heavily toward Remember-level verbs (0.479–0.600), often focusing on recall-oriented objectives. The Apply level across teaching plans appeared sporadically through verbs such as use, calculate, and solve, whereas higher-order verbs related to Analyze, Evaluate, and Create were extremely rare.

The predominance of lower-order cognitive verbs suggests that large language models tend to emulate basic textbook-style objectives that emphasize comprehension and recall. This pattern reflects the fact that the models' distribution of educational training data is dominated by instructional materials such as textbooks, worksheets, and curriculum documents that primarily use lower-order verbs emphasizing recall and comprehension. To achieve inquiry-driven or more creative outcomes, teachers need to explicitly request a mix of Bloom-level verbs and require corresponding tasks and activities to be suitable for higher cognitive levels.

### 5.6 Cross-Metric Synthesis

Integrating and summarizing all metrics reveals three dominant trends.

Firstly, model choice largely governs the readability of the teaching plan. It is shown that DeepSeek is the most accessible in terms of linguistic complexity, while the Claude and Gemini models produced teaching plans with denser academic language that is slightly beyond the reading level of grade 9-12 students.

Secondly, framework structure and model choice both affect factual accuracy to approximately the same degree. The RACE prompt framework achieved the lowest mean Hallucination Index of 2.00 and the highest NGSS curriculum alignment of 0.092. Among models, Grok was also highly reliable, averaging a 2.00 on the Hallucination Index.

Thirdly, cognitive demand stayed clustered at lower Bloom Taxonomy tiers consistently across all generated lesson plans, with all five AI models having CDI values indicating cognitive demand on the levels of Remember and Understand.

Together, these findings suggest the most effective classroom configuration pairs a readability-friendly model such as DeepSeek with the prompt framework RACE, supplemented by explicit instructions towards cognitive level, curriculum references, class content, and context.

### 6. Discussion

The results demonstrate that model type and prompt framework influence different dimensions of pedagogical quality of AI-generated lesson plans.

Readability and linguistic complexity were primarily a function of the chosen AI model. DeepSeek consistently produced plans that aligned with secondary-level reading targets, while Claude and, to a lesser extent, Grok tended toward denser academic language. Such variation in linguistic accessibility carries practical implications for classroom use, as lesson plans written at the appropriate readability level minimize the teacher's need for subsequent simplification or rephrasing.

Prompt framework exerted a stronger effect on factual completeness and curricular alignment than on readability. The prompt framework RACE reliably reduced the Hallucination Index with a mean value of 2.00 and produced the highest incidental similarity to NGSS language at the framework level with a mean similarity value of 0.082. This suggests that the procedural structure of the RACE framework, which includes both the "Role" and "Execute" sections, encourages models to generate more comprehensive and technically grounded lesson content by framing the model's output as a professional teacher's deliverable. Comparatively, the prompt framework COSTAR yielded slightly lower readability grade levels than the prompt frameworks RACE or TAG on average, but this advantage did not translate into factual accuracy or curriculum alignment. This pattern shows that stylistic guidance alone is insufficient for scientific completeness. Instead, prompt frameworks that foreground and clarify roles, actions, and execution checklists are more likely to elicit the clear physics anchors required for The Electromagnetic Spectrum.

Additionally, model reliability exhibited a narrower spread than readability but still revealed insightful differences. The AI model Grok achieved the lowest hallucination score overall, with a mean Hallucination index of 2.00, whereas DeepSeek, despite its readability advantage, omitted targeted items slightly more often, having the highest hallucination score. The absence of a consistent correlation between linguistic accessibility and reliability is notable.

The Bloom Taxonomy analysis confirmed a default lean toward lower cognitive tiers in objective statements from AI-generated lesson plans. Across AI models and prompt frameworks, Remember and Understand verbs dominated mostly, with Apply verbs appearing intermittently and Analyze, Evaluate, and Create rarely present. The average Cognitive Demand Index values between all models ranged from 1.578 to 2.294. This distribution mirrors the conventional pattern found in educational textbooks, which tend to emphasize factual recognition and conceptual explanation. Since learning objectives shape both assessment design and classroom activities, the consistent absence of higher-order verbs may constrain instructional depth, leading to lessons that prioritize recall and basic understanding over analysis or creation. Hence, with AI-generated lesson plans, educators must explicitly specify higher-order cognitive verbs to ensure that objectives extend beyond recall and comprehension.

The cross-metric synthesis points to a practical method and configuration for generating AI lesson plans that are both linguistically accessible and scientifically accurate and complete. A readability-optimized AI model, like DeepSeek, paired with the prompt framework RACE and an explicit "execute" checklist produces outputs that require less manual remediation. The checklist should include specific lesson content, explicit references to curriculum material, and inclusion of at least one Analyze-level and one Create-level objective with aligned assessments.

Several limitations qualify the interpretation of these findings. The dataset covers a single topic within high-school physics, so external validity to other topics and disciplines is not guaranteed. The sample size of fifteen plans limits the precision of dispersion estimates at both the model and framework levels. The alignment metric measures semantic similarity to NGSS curriculum descriptors rather than formal standards mapping. Finally, output quality can drift as models are updated over time, which means the relative performance observed in this study may evolve when repeated.

Despite these limitations, the convergence of results across independent metrics strengthens the trends observed. Readability differences are large and consistent enough to shape conclusions, framework effects on factual completeness are stable across models, and the low baseline for higher-order objectives is significantly evident in both the verb distributions and the cognitive demand indices. These convergences provide an actionable foundation for educators and instructional designers seeking to operationalize generative AI in planning workflows.

Future research can extend this work along several axes. Expanding to multiple physics topics and to other subjects would test the robustness of the model. Incorporating classroom-based outcomes such as student learning gains, time-on-task, and teacher editing time would connect automated metrics to instructional impact. Finally, incorporating analyses of accessibility features—such as multilingual adaptability, accommodations for learners with disabilities, and the inclusion of culturally responsive examples would expand the notion of usability beyond readability and factual accuracy, extending this research into a broader pedagogical domain.

## 7. Conclusion

This study conducted a multi-model, multi-framework evaluation of AI-generated lesson plans for The Electromagnetic Spectrum in high-school physics using four automated metrics. The evidence indicates that model selection primarily determines readability. The prompt framework structure primarily determines factual completeness and incidental curricular resemblance. Across all conditions, objectives concentrated at the lower tiers of Bloom's taxonomy.

The practical implication is clear. Educators seeking immediately usable plans should combine a readability-friendly model with a prompt framework such as RACE and supply a detailed execution checklist that specifies essential physics anchors, curriculum material, and at least one Analyze and one Create objective tied to aligned activities and assessments. This configuration minimizes post-editing, improves scientific completeness, and pushes objectives beyond recall and comprehension into higher cognitive levels.

The study's scope is limited to one topic and a modest number of plans, and its alignment and hallucination metrics are proxies rather than exhaustive audits. Even with these constraints, the patterns are clear enough to guide near-term practice and to inform future experimental designs. By pairing the right model with the right framework and by making non-negotiable pedagogical requirements explicit, AI-assisted lesson planning can produce outputs that are both classroom-ready and pedagogically sound.

## 8. Declaration of generative AI and AI-assisted technologies in the manuscript preparation process

During the preparation of this work the author used ChatGPT in order to refine Python code snippets for readability indices, verb extraction, and similarity analysis and used ChatGPT, DeepSeek, Grok, Gemini, and Claude to generate lesson plans for the aim of the study. After using this tool/service, the author reviewed and edited the content as needed and take(s) full responsibility for the content of the published article.

## 9. References

Anderson, L. W., & Krathwohl, D. R. (2001). *A taxonomy for learning, teaching, and assessing: A revision of Bloom's taxonomy of educational objectives*. Longman.

Baytak, A. (2024). The Content Analysis of the Lesson Plans Created by ChatGPT and Google Gemini. *Research in Social Sciences and Technology*, *9*(1), 329-350. https://doi.org/10.46303/ressat.2024.19

Dornburg, A., & Davin, K. J. (2025). ChatGPT in foreign language lesson plan creation: Trends, variability, and historical biases. *ReCALL*, *37*(3), 332–347. doi:10.1017/S0958344024000272

Dornburg, A., Davin, K. (2024). To what extent is ChatGPT useful for language teacher lesson plan creation?. arXiv preprint arXiv:2407.09974.

Fan, H., Chen, G., Wang, X., Peng, Z. (2024). LessonPlanner: Assisting Novice Teachers to Prepare Pedagogy-Driven Lesson Plans with Large Language Models. arXiv preprint arXiv:2408.01102.

Flesch, R. (1948). A new readability yardstick. *Journal of Applied Psychology*, *32*(3), 221–233. https://doi.org/10.1037/h0057532

*HS-PS4 Waves and their Applications in Technologies for Information Transfer | Next Generation Science Standards*. (n.d.). www.nextgenscience.org. Retrieved October 11, 2025, from https://www.nextgenscience.org/dci-arrangement/hs-ps4-waves-and-their-applications-technologies-information-transfer

Kahraman, N., & Kıyıcı, G. (2025). Evaluating the Efficacy of AI-Generated Inquiry-Based Lesson Plans in Science. Sakarya University Journal of Education, 15(1), 40-53. https://doi.org/10.19126/suje.1463067

Karaman, M. R., & Göksu, İ. (2024). Are Lesson Plans Created by ChatGPT More Effective? An Experimental Study. International Journal of Technology in Education, 7(1), 107–127. https://doi.org/10.46328/ijte.607

Lee, G., Zhai, X. (2024). Using ChatGPT for Science Learning: A Study on Pre-service Teachers' Lesson Planning. arXiv preprint arXiv:2402.01674.

Powell, W., & Courchesne, S. (2024). Opportunities and risks involved in using ChatGPT to create first grade science lesson plans. *PloS One*, *19*(6), e0305337–e0305337. https://doi.org/10.1371/journal.pone.0305337

Thomas, G., Hartley, R. D., & Kincaid, J. P. (1975). Test-Retest and Inter-Analyst Reliability of the Automated Readability Index, Flesch Reading Ease Score, and the Fog Count. *Journal of Reading Behavior*, *7*(2), 149–154. https://doi.org/10.1080/10862967509547131

## 10. Appendix

**Appendix A**
Due to length, all supplementary materials (full lesson-plan outputs, code, and metric summaries) are hosted externally at:
https://drive.google.com/drive/folders/12xupOFLrIV1s-yOFDkWKaPQUXK2IjG0J?usp=sharing .